# EMOTHAW: A novel database for emotional state recognition from handwriting

Laurence Likforman-Sulem [a] Anna Esposito [b] Marcos Faundez-Zanuy [c]

Stéphan Clémençon [a] and Gennaro Cordasco [b]

[a]*LTCI, CNRS, Télécom ParisTech, Université Paris-Saclay, Paris, France*
[b]*Second University of Naples, Caserta and International Institute for Advanced Scientific Studies-IIASS, Salerno, Italy*
[c]*Escola Universitaria Politecnica de Mataro, TecnoCampus Mataro-Maresme, Spain*

**Abstract**

The detection of negative emotions through daily activities such as handwriting is useful for promoting well-being. The spread of human-machine interfaces such as tablets makes the collection of handwriting samples easier. In this context, we present a first publicly available handwriting database which relates emotional states to handwriting, that we call EMOTHAW. This database includes samples of 129 participants whose emotional states, namely anxiety, depression and stress, are assessed by the Depression Anxiety Stress Scales (DASS) questionnaire. Seven tasks are recorded through a digitizing tablet: pentagons and house drawing, words copied in handprint, circles and clock drawing, and one sentence copied in cursive writing. Records consist in pen positions, on-paper and in-air, time stamp, pressure, pen azimuth and altitude. We report our analysis on this database. From collected data, we first compute measurements related to timing and ductus. We compute separate measurements according to the position of the writing device: on paper or in-air. We analyse and classify this set of measurements (referred to as features) using a random forest approach. This latter is a machine learning method [2], based on an ensemble of decision trees, which includes a feature ranking process. We use this ranking process to identify the features which best reveal a targeted emotional state.

We then build random forest classifiers associated to each emotional state. Our results, obtained from cross-validation experiments, show that the targeted emotional states can be identified with accuracies ranging from 60% to 71%.

## 1 Introduction

Health care mainly relies on the early detection of illnesses, and clinical tests have been developed to diagnose diseases and follow their evolution. Among tests, those based on human activity (speech, handwriting, body movements) have the advantage of being non invasive and are valuable tools for complementing laboratory analyses and clinical examination. In particular, simple pen and paper tests can detect cognitive impairment through handwriting: lack of legibility, jagging and perseveration of letters are well-known effects of Parkinson (PD) and Alzheimer (AD) diseases [22].

The importance of detecting early signs of illnesses can be extended to the detection of negative emotions since emotions such as depression, anxiety and stress influence health. Depression is a complex and heterogeneous mood disorder that is expressed by behavioral disinterest and sad feelings and it may cause serious social, occupational and cognitive impairments [33] [29]. As defined by Eysenck et al. [6](pp.336) *"Anxiety is an aversive emotional and motivational state occurring in threatening circumstances"*. It affects cognition and reduce the individual's effectiveness and efficiency in performing cognitive tasks. The most comprehensive definition of stress is *"a [negative] emotional experience accompanied by predictable biochemical, physiological and behavioral changes"* cited in [1]. The causes of stress are extremely diverse, ranging from difficulties to handle everyday experiences and changes to traumatic events such as surviving to a natural disaster. Depression, anxiety and stress, are natural responses to changes and challenges of everyday life. However, when persisting over a long time period, they can produce serious illnesses such as Major Depression Disorders (MDD), Generalized Anxiety Disorder (GAD) and Chronic Stress [29] [3] [20].

Mundt et al. [21] and Yang [37] have exploited the speech signal to detect depression. They report significant differences in the acoustic biomarker values (such as F0 and F0-derived measures) of clinical/non-clinical subjects. However, as pointed out in Esposito & Esposito [5] *"[voice acoustic measures] appear to be affected at various degrees by many sources of variability that causes distortions and modifications in the original signal, thus modifying the acoustic features useful for its [automatic] recognition"*. In handwriting acquisition however, the signal is easily acquired without error propagation.

In the present study we propose to detect negative emotions such as depression, anxiety and stress through handwriting, a human daily activity. Our approach consists in collecting an individual's handwriting through a computerized platform and predict his/her emotional state through a machine-learning approach.

Machine learning approaches such as Support Vector Machines, Neural Networks, Bayesian Networks have been used successfully in related domains such as affective computing and personality computing [34] where measurements (referred to as features) are extracted from behavioral signals such as face expressions, speech or body gestures, and fed to a classifier. The outcome of the classification in affective computing is one simple emotion among the set of basic ones: happiness, surprise, sadness, anger, fear, disgust. The goal is to improve human-machine interfaces by adequately reacting from users' inputs. For personality computing, the outcomes are personality traits: openness, agreeableness, conscientiousness, extraversion, neuroticism. Machine-learning approaches require labeled databases for training the classifier [10]. The personality trait labels are most often assessed through the BigFive questionnaire [11]. Similarly to BigFive and personality, negative emotions such as depression, anxiety and stress can be scored through the DASS (Depression-Anxiety-Stress Scales) scales [18]. These scales are now well assessed [4] and also use a self-reported questionnaire. Administering the DASS questionnaire to each participant, we have built "EMOTHAW" (EMOTion recognition for HAndWriting) a first publicly available database [1] relating emotional states to handwriting. Other handwriting databases devoted to biometry [8] or recognition [12] have been developed, and do not include emotional labels.

In addition to the database, we present a non-parametric classifier based on the random forests machine learning approach [2]. We extract features related to timing and ductus from the collected data. We compute separate features according to the position of the writing device: on paper or in-air. Random forest training provides ranking of the input features according to their importance. We develop an analysis of the extracted features based on these rankings. From this analysis, we deduce which tasks and which features better characterize a targeted emotional state. We also build random forest classifiers based on the extracted features, and provide recognition results for the three emotional states covered by the DASS scales.

Our paper is organized as follows. In Section 2, we present previous computerized studies devoted to handwriting analysis for diagnosis purposes. We formulate the assumptions of our analysis in Section 3. Section 4 describes the process of data collection as well as the cohort recruitment. The set of features extracted on raw handwriting data is described in Section 5.1. Section 5.2 presents our feature ranking and classification of emotions, based on random forests. In Section 6, we apply this analysis to the collected data and provide recognition results.

---

[1] the database will be publicly available by the time of publication

## 2  Handwriting analysis for diagnosis purposes

Handwriting collected on paper has long been used as a means for authentication, cognitive impairment detection and personality trait assessment. Indeed, collecting handwriting is not invasive, simple and cheap and requires little expertise from the operator. Various pen and paper tests have been developed and used to complement laboratory data, physician examination or face-to-face interviews.

From handwritten samples, it has been observed that Alzheimer and Parkinson patients [7] deteriorate character shapes, add or omit characters within words and have less fluent movements. In the biometry domain, off-line signatures have long been used as a means of authentication as well as writing samples collected with or without the knowledge of the writer [24]. Handwriting analysis have also lead to graphology-related applications such as detection of lies. In this context, Tang [36] assumes that when honest people lie, their writing is modified due to the associated cognitive stress. Writing may be less fluent and margins may be modified at places where the lie is expressed. Well-assessed tests in the medical domain are:

- The clock drawing test (CDT) where participants are required to draw a clock, including all digits, and setting the hands to 11:50. Patients with Alzheimer disease may not space digits evenly and sometimes make mistakes when setting the clock hands. A maximum score of 15 points is awarded for shape and spatial arrangement of the clock numbers, and hands.
- The MMSE test (Mini Mental State Examination) [9] is a 30-point test that evaluates cognitive functions related to registration, attention and calculation, recall, language, ability to follow simple commands and orientation. Among other tasks, the subject is asked to write a sentence of his own, draw interlinking pentagons and memorizing a sequence of 3 words.

- The HTP (House-Tree-Person) test, designed by J. Buck [16] is a clinical test where participants are asked to draw a house, a tree and a person. The drawing tasks are accompanied by questions to infer personality. This test is sometimes extended to evaluate brain damage.

Due to the development of scanners and tablets, the following studies aim at converting pen and paper tests to computerized platforms. The collected data within computerized platforms are on-line in contrast to off-line. The advantage of on-line data is that non-directly visible features such as speed, pressure, number of strokes, stylus inclination, etc... are recorded. It is sometimes possible to record both on-line and off-line data using a special ink-based stylus. On-line signatures include in addition to pen positions, pressure, stylus inclination and azimuth which cannot be observed in off-line signatures [25]. The CDT test has been computerized into the ClockMe system [15] which automatically evaluates the CDT score, following the paper-based protocol. No difference can be observed between the paper and tablet based scores, the tablet offering supplemental data awaiting to be processed and interpreted.

Recently, the 'strokes against stroke' system has been proposed to detect brain stroke risk from stroke measurement achieved on a tablet from stimuli displayed on a screen [27]. A previous version of this system [23] was based on signatures and triangle drawings.

The CompPET system [13] (Computerized Penmanship Evaluation Tool) is a computerized platform which collects on-line handwritten data, including pressure, in air and on tablet points. From data collected with ComPET, changes in handwriting have been measured according to age, and various major health disorders such as depression (MDD: Major Disorder Depression). Recently, this platform has been used to automatically detect lies in handwriting [19], using speed and size of handwriting units.

The previous studies deal with handwriting analysis but none of them deal with emotion recognition. Our present contribution provides a handwriting database and an analysis of that database relating emotional states to handwriting tasks and pen movements.

## 3 Setup for our analysis

We assume that handwriting is related to behavior, and it is influenced by one individual's emotional state. We focus here on negative emotional states such as depression, anxiety and stress, which are seen as distinct states. However, their clinical symptoms largely overlap [18]. The DASS scales (Depression-Anxiety-Stress Scales) proposed by Lovibond and Lovibond (1995) [18] have been designed to provide a maximum discrimination between these negative emotional states. The items included in each scale have been chosen in order to produce three orthogonal axes from a factor analysis: DASS-Depression focuses on items related to low motivation and self esteem, DASS-anxiety to fear and perceived panic, and DASS-stress to tension and irritability.

Following this classification, we assume that each emotion can be separately recognized through measurements on handwriting, and that each emotional state is likely to be related to specific writing tasks and measurements.

Starting with little knowledge on which state influences which measurement, we propose a machine learning approach, namely random forests [2], which automatically ranks the measurements associated to each recognition task. The random forest approach also provides an estimate on the participants' emotional state from these measurements.

The starting point of this study is the collection of a database from 129 participants, which we refer to as the EMOTHAW database. On-line data are collected through a computerized platform. This database is in itself a useful tool due to the lack of publicly available labeled data for this domain. The writing/drawing tasks from which the measurements are extracted are well assessed tasks used in medical diagnosis or in scoring handwrit-ing proficiency (house drawing, text copying, ...). The emotional state of participants is assessed by the DASS scale. From the database, 20 measurements are extracted, related to the kinematic and ductus of the writing and likely to change under the mentioned emotional states. These measurements (referred to as *features* in the machine learning terminology) are fed to classifiers in order to automatically recognize the above emotional states.

## 4   Data collection

We describe here the data collection process of the EMOTHAW database: raw measurements provided by the digitizing tablet, tasks completed by the participants, and ground truth assessment.

### 4.1   Computerized platform

Handwriting data have been registered thanks to an INTUOS WACOM series 4 digitizing tablet and a special writing device named Intuos Inkpen. This device provides high spatial and pressure accuracies and can be considered as a state-of-the-art tablet. Participants were required to write on a sheet of paper (DIN A4 normal paper) laid on the tablet. Figure 1 shows the sample acquired from one user. No time restrictions were provided to the user. While the digital signal is visualized on the screen it is also visible in the paper (due to the inkpen), and the user normally looks to the paper. There was a human supervisor sat next to him but he did not take care of this task on the paper. He just controlled the computer. It was the same procedure that was used in the MCYT and BIOSECURID databases [8].

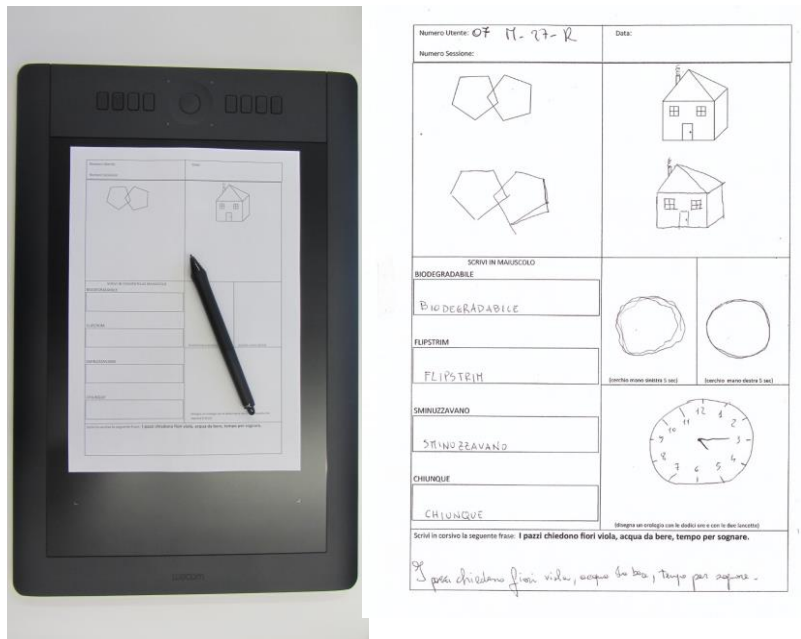

Fig. 1. left: Acquisition tablet with ready to fill A4 sheet and writing device. right: sheet with the whole set of tasks filled by one user.

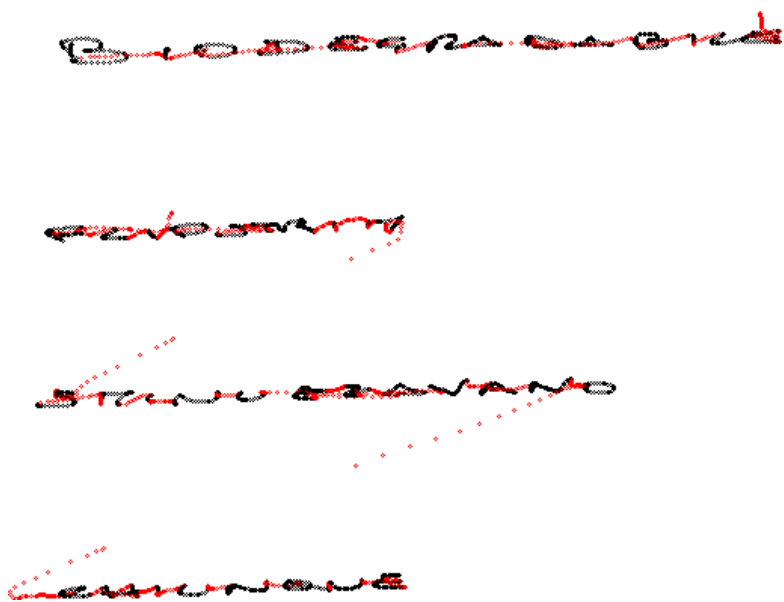

Fig. 2. Task 3 sample with on-paper points (in black, pen status down=1) and in-air points (in red, pen status up=0). Additional in-air movements (writing device too high above the tablet) are not registered.

The tablet acquires the task in real time. We have specially developed an acquisition software to store the tasks and users in folders in the hard disk. The resulting files are svc files, svc being the file extension provided by Wacom. Svc files are ASCII files that can be opened with WORD, NOTEPAD and other standard editor applications.

The following information is captured:

(i) Position in x-axis.

(ii) Position in y-axis.

(iii) Time stamp

(iv) Pen status (up=0 or down =1)

(v) Azimuth angle of the pen with respect to the tablet (see Fig. 4).

(vi) Altitude angle of the pen with respect to the tablet (see Fig. 4).

(vii) Pressure applied by the pen.

Using this set of dynamic data, further information can be inferred, such as velocity features (acceleration, velocity, instantaneous trajectory angle, instantaneous displacement, etc...) or time features, ductus-based features (see Section 5.1). In addition, the system has the nice property to acquire in-air movements, which are lost using the on-paper ink. The in-air information has proved to be as important as the on-surface information [35] [31]. However when the writing device is too far from the tablet, in-air points are not registered. Figure 2 shows pen-down and pen-up points acquired at 100 points per second, corresponding to handwriting in capital letters acquired at 100 points per second. When the speed of writing is low, registered points are close to each other and strokes seem darker than when the speed is high. If the stylus is risen too far from the tablet the signal is lost and no coordinates are recorded by the software. In our case for each sample we have used a vector of seven parameters, which consists of the previous list of measurements provided by the WACOM pen tablet. Figure 3 shows a short extract of an svc file. The extract includes both on-paper points (pen status equal to 1) and in-air points (pen status equal to 0). It can be noted that timestamp values always increase and that pressure (column 7) is equal to zero for in-air points. For instance, the point drawn at time stamp 17606786 is an in-air point at x position 50621 and y position 33860. Its pen status and pressure are both equal to 0, and azimuth and altitude values are equal to 1900 and 540, respectively. Azimuth and altitude values can be normalized using maximum values as normalization factors. The normalized values are in degrees: azimuth=1900 $\times$ 360/4095 = 167° and altitude=540 $\times$ 90/1023 = 47.5°.

```
                y position
      1796                                        altitude
49076  34584   17606448  1  1870  560  45
49025  34608   17606456  1  1870  560  81    azimuth
49009  34613   17606463  1  1870  560  157
x position
48995  34614   17606478  1  1870  560  193
48993  34614   17606486  1  1870  560  219   pen status: on paper
48993  34614   17606493  1  1860  560  246
48993  34614   17606501  1  1860  550  284
...   ....   ....        ...   ....   ....
50786  33795   17606756  1  1900  550  305
50727  33808   17606764  1  1900  540  130   pen status: in air
50727  33808   17606771  0  1900  540  0
50640  33840   17606779  0  1900  540  0     time stamp
50621  33860   17606786  0  1900  540  0
50619  33878   17606794  0  1900  540  0
...   ....   ....        ...   ....   ....
51032  33781   17607320  0  1940  510  0
51032  33781   17607328  1  1940  510  84    pressure
51056  33773   17607336  1  1940  510  118
...   ....   ....        ...   ....   ....
```

Fig. 3. Extract of an svc file corresponding to the pentagon drawing task. The file includes 1796 points, each one having 7 measurements (x position, y position, time stamp, pen status, azimuth, altitude and pressure). The extract includes both on-paper points (pen status equal to 1) and in-air points (pen status equal to 0).

The tasks acquired by the tablet are the following:

I. Copy of a two-pentagon drawing

II. Copy of a house drawing

III. Writing of four Italian words in capital letters (BIODEGRADABILE (biodegrad-

able), FLIPSTRIM (flipstrim), SMINUZZAVANO (to crumble), CHIUNQUE (anyone))

IV. Loops with left hand

V. Loops with right hand

VI. Clock drawing test

VII. Writing of the following phonetically complete Italian sentence in cursive letters (I pazzi chiedono fiori viola, acqua da bere, tempo per sognare *Crazy people are seeking for purple flowers, drinking water and dreaming time*).

The two-pentagon drawing is part of the minimental test (MMSE test), while drawing a clock is part of the Clock Drawing Test (CDT). House drawing is part of the HTE (House-Tree-Person) projective test of personality (see Section 2). Copying text has been also used for assessing writing abilities [30]. Thus two of the proposed tasks consist in copying a sequence of words, according to two types of script: cursive and handprinted.

The handprinted words (Biodegradabile,...) have been selected based on our previous Spanish database BIOSECURID [8]. They are neutral words that do not have positive or negative connotations. The cursive sentence (I pazzi....) is a phonetically complete word sequence which was reviewed by PhD experts at the Psychology department of the Seconda Universita di Napoli. No priming effect [36] related to the meaning of the sentence was considered here, though a possible negative effect could be related to the word "pazzi' which means "fool". But in italian the word "pazzi" is commonly used for kidding, poetry and in the daily conversational language as synonym of "special persons" particularly when they were asking for "purple flowers, drinking water, and dreaming time" (lyrics of a famous song from F. de Gregori intitled "I matti" ) . Thus, in Italian the connotation of this sentence is smoother than it could be in another language.

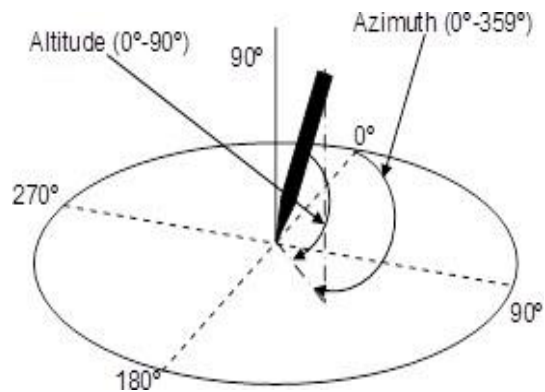

Fig. 4. Azimuth and inclination angles of the pen with respect to the plane of the graphic card.

*4.2 Participants and Emotion Scoring*

For the data collection, 129 subjects aged between 21 and 32 years, mean age 24.8 years, standard deviation SD= 2.4 years) were recruited, all Master and BS students at the Seconda Universit di Napoli, Department of Psychology, located in Caserta, Italy. It can be noted that in medical and psychological studies a database of 129 subjects with a specific background and age is a noteworthy research accomplishment. In these fields, acquiring a large database is not simple and even case studies are of great interest. There were 71 Female and 58 Male participants. The range of years has been limited in order to reduce the inter-subject variability of the experiment. Otherwise it would be difficult to attribute differences to years, or psychological conditions. There can be differences in handwriting features because of ages also and this would have been interfered with differences due to the emotional states. Each participant first filled in and signed a consent form providing her/his general demographic information. Participants' emotional state was determined by using the Italian version of the Depression Anxiety Stress Scales (I-DASS-42). The construct validity of DASS-42 was assessed by Henry and Crawford in 2003 [4] for the English version and by Severino (http://www2.psy.unsw.edu.au/dass/Italian/Severino.htm) for the Italian one. The I-DASS-42 is a 42-item questionnaire including three self-report scales, each containing 14 items, designed to measure depression, anxiety and stress. For

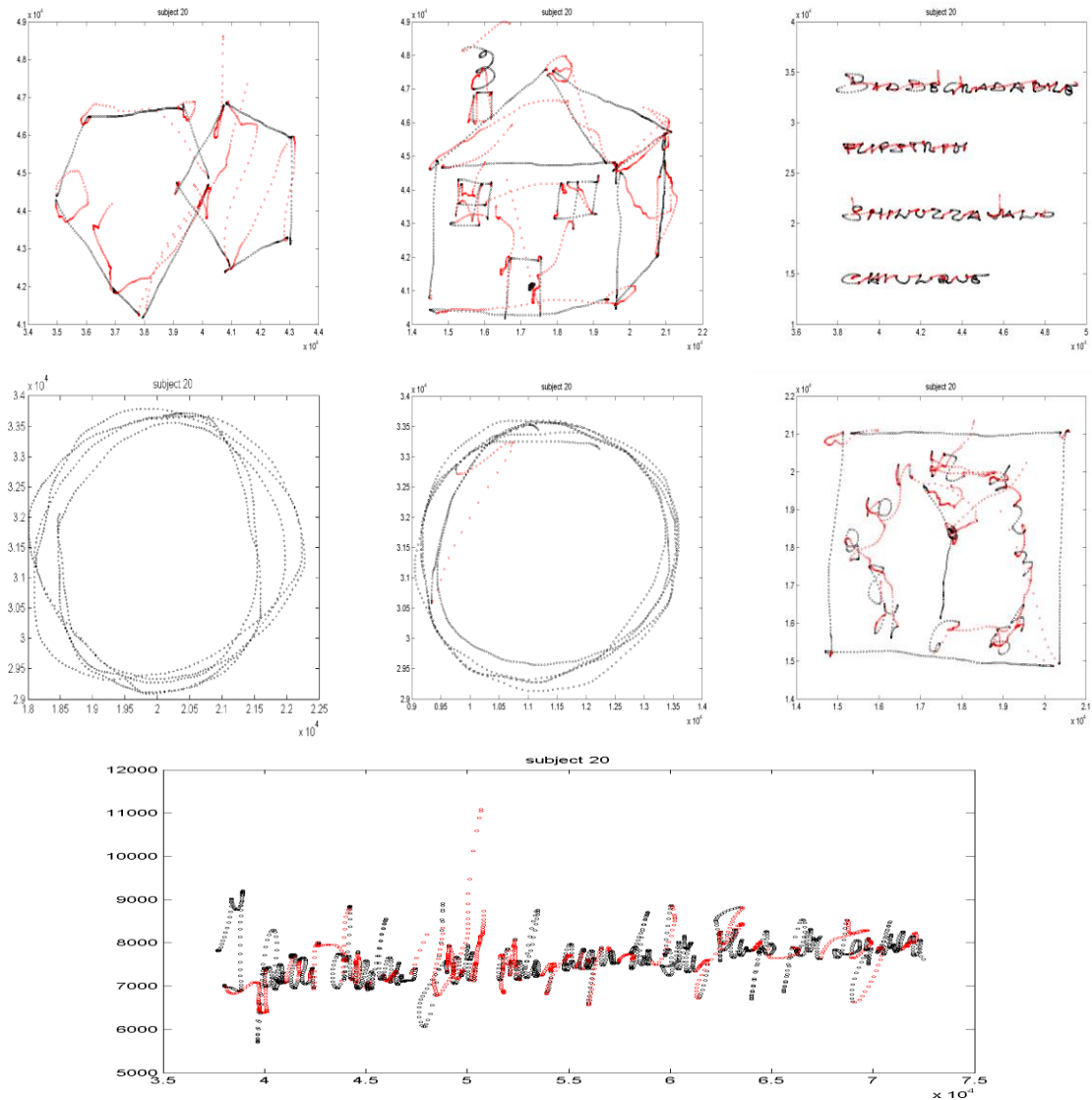

Fig. 5. Handwriting samples collected from all tasks: pentagons and house drawings, handprinted writing, loops (left hand and right hand), and clock drawing and cursive writing. Pen-down and pen-up data points are in black and red respectively.

each of these negative emotional states, subjects are asked to rate on a 4-point severity/frequency scales (from 0 to 3) the extent to which they have experienced each item [2] listed in the questionnaire over the past week. The severity-rating index assigned to each subject is made according to Table 1. The distributions of DASS scores in the EMOTHAW

[2] such items include : "I found myself getting upset rather easily", and "I had a feeling of shakiness (eg, legs going to give way)"

database are shown in Fig. 6.Likert

|  | Depression | Anxiety | Stress |
| --- | --- | --- | --- |
| Normal | 0-9 | 0-7 | 0-14 |
| Mild | 10-13 | 8 - 9 | 15 - 18 |
| Moderate | 14 - 20 | 10 - 14 | 19 - 25 |
| Severe | 21 - 27 | 15 - 19 | 26 - 33 |
| Extremely Severe | 28+ | 20+ | 34 + |

Table 1

DASS score range according to emotional state level.

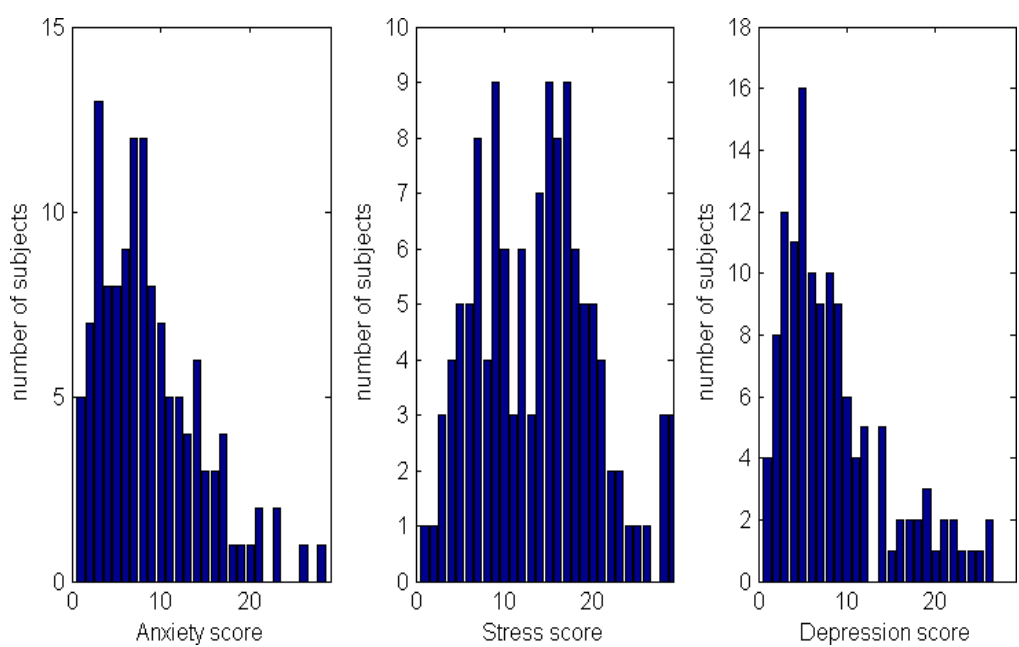

Fig. 6. DASS score distributions in the EMOTHAW database.

The English, as well as, the Italian version of the DASS can be downloaded from the website http://www2.psy.unsw.edu.au/dass/. Scoring details are described in the document found at http://www.aasw.asn.au/document/item/2794. It can be noted that DASS questionnaire is intended to screen normal adolescents and adults for detecting psychological mood disorders: severe disorders should be assessed through clinical examination. Cross-tables in Fig. 7 show that depression, anxiety and stress can be observed separately or in conjunction. The scores have been dichotomized as explained in the following (see Section 5). From the matrixes in the 2nd row of Fig. 7, we observe that for about 20% of the participants, a single negative state is observed (f.i. such as anxious/non stressed/non depressed). For about the same percentage of participants (22%), two negative emotional states are observed in conjunction. However, due to the scales' construct, each of these emotional states can be predicted separately. A Pearson's $\chi^2$ test conducted on the anxious/stressed, stressed/depressed and anxious/depressed cross-tables, shows that the qualitative variables anxiety and stress, as well as stress and depression and depression and anxiety, are linked (p-values below 0.01). The strongest link is between anxiety and stress. Then the link is stronger between anxiety and depression, than between stress and depression.

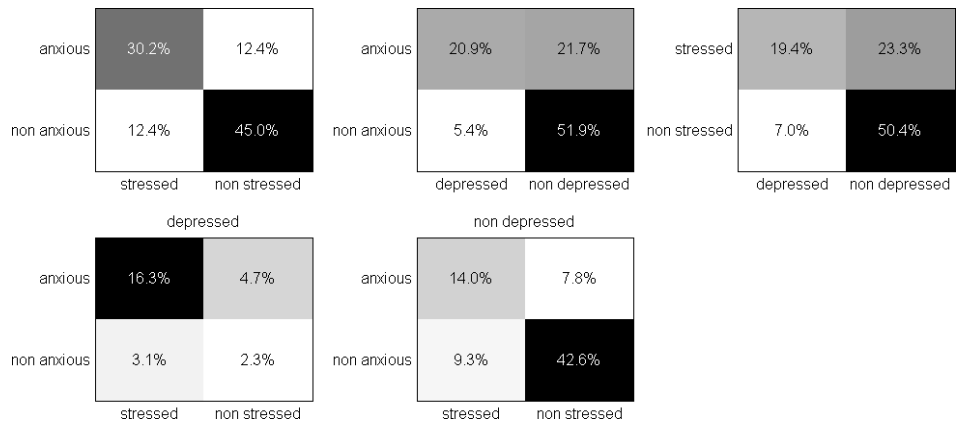

Fig. 7. Cross tables showing the percentage of co-occurrence of emotional states in the EMOTHAW database.

## 5  Emotional State Recognition

In the following experiments, DASS scores have been dichotomized in order to predict the state of an individual as a two-class problem: anxious or not anxious, stressed or not, or depressed or not. This is motivated by the fact that approaches such as ordinal regression would need more sample points to build the regression function. The non anxious state refers to normal anxiety level (score less than 7), while the anxious state refers to mild to high anxious levels (scores up to 20). The stressed emotional state is scored more than 14 and the depression state, more than 9 (see Table 1).

### *5.1  Proposed features and tasks*

Feature extraction is the first step of an automatic recognition system. It consists in representing the raw data signal in a more concise and accurate way. Indeed, the registered handwriting signal includes raw features such as x-y positions of the writing device, as well as absolute time and pressure at each of x-y position (see Section 4.1). From these raw features, accurate features must be extracted. Such features should be independent of absolute pen position and time, as well as being efficient in solving the targeted state recognition task.

The definition of features for recognizing emotional states from handwriting is not straightforward. To our knowledge, there is no assessed features for such tasks. Thus our approach consists in proposing a number of features, extracted from the tasks described above (see Section 4), then ranking and analyzing their importance through a random forest approach (see Section 5.2).

The first features we propose are timing-based features which have proven to be efficient for assessing writing proficiency of pupils and elderly people copying texts [31] [32]. Both time spent on paper and in air were found efficient. We extend such timing-based features to cursive writing and drawing tasks. We have also observed that ductus (ductus is the way how strokes are drawn including stroke order, direction and speed) presents great variations across data. Ductus may be related to individual differences but it may also be related to the emotional state of an individual. We have thus added a feature related to ductus. For each task, we thus propose the following timing-based and ductus-based features:

- $F_1$: time spent in-air while completing the task
- $F_2$: time spent on-paper while completing the task
- $F_3$: time to complete the whole task
- $F_4$: number of on-paper strokes while completing the task

A stroke in this context corresponds to consecutive drawing points achieved without lifting the pen. From the seven registered tasks, the two loop drawings have been removed since such tasks do not include any pen-up movement. Thus we extract these four features on five tasks, cumulating 20 measurements for each subject. In the following, we analyse these measurements by a machine-learning approach in order to identify those that are important for a targeted emotional state.

### 5.2 Feature analysis with random forests

We have proposed above (see Section 5.1) a set of 20 measurements, referred to as features, to extract from five handwriting/drawing tasks. Our objective is now to find which tasks and which features are relevant in order to recognize an emotional state. A machine-learning approach, namely random forests [2], satisfies this objective by providing a set of importance measures for each feature along with training.

Training a random forest consists in building $T$ decision trees which are combined at decision level. Each tree is built from a subset of the training data and from a subset of the features or variables. A random process occurs for the selection of data used for training the $t^{th}$ tree and another one for the selection of features at each split node of this tree. At each node, the candidate set of features used for splitting the node into child nodes, is restricted to *mtry* features. This restricted set is obtained by randomly sampling the original feature set. The training data, not selected by the first random process, are used for testing the efficiency of this tree. Such data are the so-called OOB (Out-Of-Bag) data [17]. Since each data point may be an OOB for several trees in the forest, all decisions for this OOB can be recorded and accumulated. The final decision for an OOB is provided by a majority vote. OOB decisions permit to compute the so-called OOB error rate which is a valuable estimation of the forest accuracy.

Another advantage of the random forest approach is that variable relative importance measures are provided through the training process. There are four importance measures for a given feature $f_i$:

- Measure 1: is the amount of decrease of the OOB error rate. It is computed as follows. The value of feature $f_i$ is randomly changed for each out-of-bag data point. The error rate is thus modified compared to using the original data. If the error rate significantly increases, this means that feature $f_i$ is important. Conversely if it decreases, the importance value is negative and this means that feature $f_i$ is not reliable.
- Measure 2: average margin decrease. The so-called margin of an OOB point is the difference between the proportion of trees in the forest which classify correctly this OOB point minus the proportion of trees which misclassify this point. When the value of feature $f_i$ is randomly changed, the margin is modified (it typically decreases). The decreases are averaged across OOB points to provide measure 2.
- Measure 3: is the normalized difference between the number of margins which have

decreased and the number of margins which have increased when applying the previous process ($f_i$ randomly changed for OOBs).

- Measure 4: mean decrease of the Gini criterion. The Gini criterion is an impurity measure used to choose the best split variable at each tree node. For a given feature $f_i$, one can sum up the gains in impurity obtained when using this feature in the forest trees. This sum is then normalized by the number of trees in the forest to provide measure 4 for $f_i$.

Features are then ranked according to each importance measure and each targeted emotional state. The automatic ranking process is the following. For a given emotional state, the ranks of each feature according to each importance measure are summed up. The lowest the sum, the best the feature. However, since building a forest includes a part of randomness, the ranking of features may slightly vary from one forest to another. Thus, an ensemble of $T$ forests is built and the ranks are cumulated from each forest of the ensemble.

Random forests are mostly used for classification purposes. Thus, we also train three random forest classifiers from the previous set of measurements in order to recognize depression, anxiety and stress.

# 6 Experiments

We apply our analysis based on random forests (see Section 5) to the recognition of anxiety, depression and stress.

## 6.1 Feature analysis

The first analysis consists in ranking the proposed timing and ductus-based features (see Section 5.1). The process starts by building a random forest ensemble from the set of extracted features (20 features: 5 tasks, 4 features per task). For each emotion, a forest is built according to Section 5.2, using *ntree* trees, and at each tree node, *mtry* variables are considered for a split. In this study, forests are built with a large number of trees (*ntree*=100). Since we have *nfeat* = 20 features, *mtry* = $\sqrt{nfeat}$ ∼ 5 is a popular tuning for the number of variables considered at each tree node.

Random forest training provides both a forest of *ntree* trees and four importance measures for each feature (variable) considered. Features are ranked according to the automatic ranking process described above, using an ensemble of *N* = 50 forests (see Section 5.2). The ten top-ranked features provided by each random forest model are shown in Table 2.

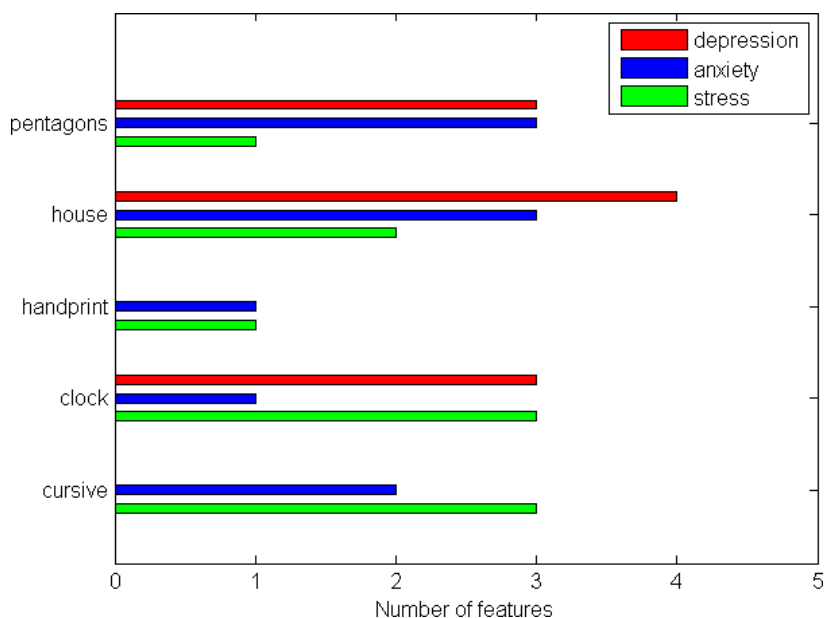

Fig. 8. Number of features according to writing/drawing tasks and emotional state.

We observe from Fig. 8 and Table 2 that depression is expressed through drawing tasks

| Random Forest Model | Features |
|---|---|
| Depression | in-air duration (clock), on-paper duration (clock) |
| | total duration (clock), in-air duration (pentagons), |
| | total duration (pentagons), in-air duration (house) |
| | on-paper duration (pentagons), total duration (house) |
| | on-paper duration (house), number of pen-down strokes (house) |
| Anxiety | on-paper duration (clock), total duration (cursive) |
| | on paper duration (pentagons), in air duration (cursive) |
| | in air duration (house), number of pen-down strokes (house) |
| | in air duration (pentagons), total duration (handprint) |
| | total duration (house), total duration (pentagons) |
| Stress | on paper duration (clock), in air duration (cursive) |
| | total duration (clock), on-paper duration (pentagons) |
| | in-air duration (clock) , number of pen-down strokes (cursive) |
| | on-paper duration (house), on-paper duration (handprint) |
| | total duration (cursive), in air duration (house) |

Table 2

Top ranked features according to each model.

only (clock, pentagons, house). In contrast, anxiety and stress use both handwriting (printed, handprinted) and drawing cues.

We also observe that both features related to in-air and on-paper motions are useful for characterizing emotions. In Table 2, all models use three in-air cues (in air duration) picked from the proposed handwriting/drawing tasks. This shows the importance of in-air cues which cannot be observed in the ink trace. In contrast, the proposed ductus-related cues (number of pen-down strokes) are used only once for each model.

*6.2 Emotion recognition*

To provide emotion recognition results, we conduct $C = 10$ repetitions of leave-one-out cross validation experiments with $K = 129$ folds. Each one-leave-out cross validation experiment consists in building a forest from all but one data point (the $K - 1$ folds), then using this forest to test the remaining data point (the $K$th fold). This process is repeated $K$ times. From the $C$ repetitions of this process, $C$ cross validation results are collected and the mean accuracies are shown in Table 3 for each model. We also provide these results through box-and-whisker plots (see Fig. 9). All random forest experiments have been conducted using the R language, and the random forest and ipred packages [28].

| Random Forest Model | Accuracy (in %) |
|---|---|
| Depression | 71.2 |
| Anxiety | 60 |
| Stress | 60.2 |

Table 3

Mean recognition accuracies for each model, provided by $C = 10$ repetitions of leave-one-out cross-validation experiments.

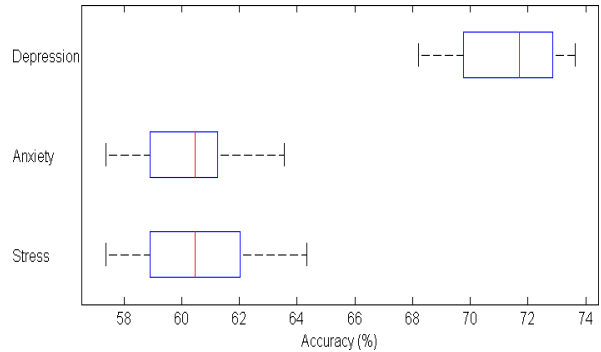

Fig. 9. Performance of emotional state recognition displayed through box-and-whisker plots.

Depression recognition performs the best among the proposed recognition tasks. Mean performance accuracies for anxiety and stress recognition are 60% and 60.2 % respectively, compared to 71.2% for depression recognition (see Table 3). However anxiety and stress accuracies are both above chance level.

Fig. 10 shows clock drawings from several participants: participant 48 is normal (non-depressed, non stressed), participant 49 is depressed 49 and participant 19, stressed. The duration on paper is much larger for the depressed participant (36,8 seconds) than for the non-depressed one (8,7 seconds). This is also true for the duration in-air: longer for the depressed participant (39 seconds) than for the non-depressed one (22 seconds). This can be related to studies on speech processing where it has been shown that speech pauses are longer for depressed participants. The in air duration is also higher for the stressed participant of Fig. 10-c (46s), compared to the non stressed one (22s). From the red points of Fig. 10-b and c, we observe that in-air movements of stressed and depressed participants differ: they are erratic for the stressed participant, smooth for the depressed one. Thus a long in air duration is indeed the sign of a disorder but more features are necessary to distinguish between a stressed participant and a depressed one. This shows the interest of using several features for emotion recognition.

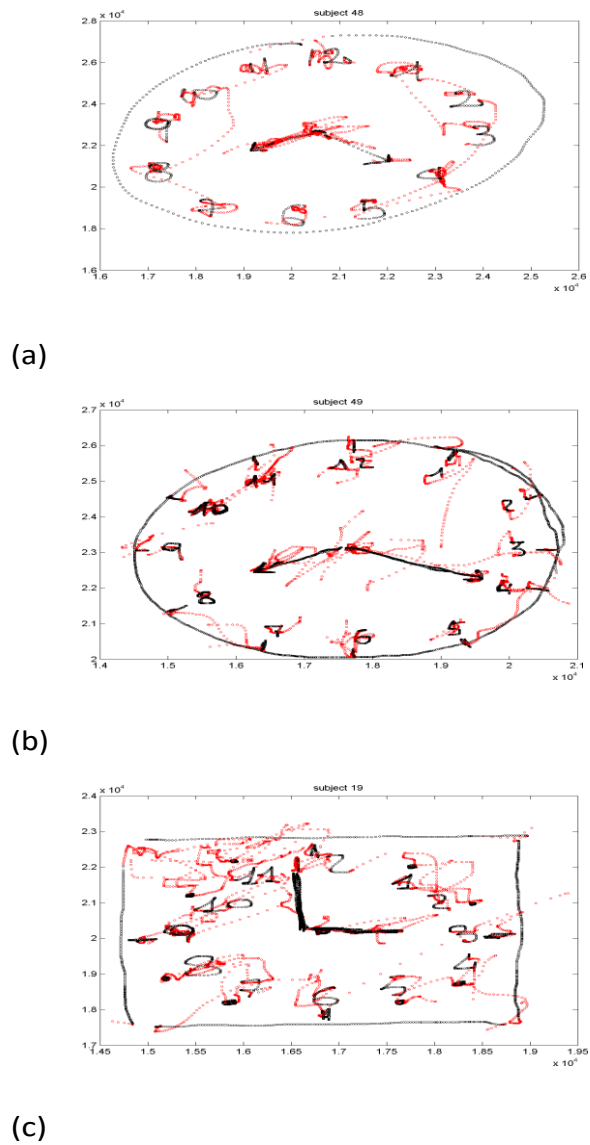

(a)

(b)

(c)

Fig. 10. Sample clock drawings. a) non stressed/non depressed participant 48: DASS-depression 3, DASS-anxiety 6, DASS-stress 6. b) depressed participant 49: DASS-depression 11, DASS-anxiety 2, DASS-stress 5. c) stressed participant 19: DASS-depression 9, DASS anxiety 4, DASS-stress 15.

## 7 Conclusion

We have presented EMOTHAW, a novel database devoted to a new task: the recognition of emotional states from handwriting. The affective database includes handwritten samples from 129 participants whose emotional states, namely depression, anxiety and stress, are assessed by the DASS scales. Seven tasks are recorded through a digitizing tablet: pentagons and house drawing, words copied in handprint, circles and clock drawing, and one sentence copied in cursive writing. Records consist in pen positions, time stamp, pressure, pen azimuth and altitude. These samples were registered as on-line data including both on paper and in air points. Each participant also completed the self-reported DASS questionnaire from which DASS scores were computed.

Recognition experiments have been conducted using the random forests machine learning approach. In this context, we have computed features related to timing both from in air and on paper movements, and ductus (number of strokes) . We have developed an analysis based on random forests in order to highlight important features associated to a targeted emotional state. Our results show that both in-air and on-paper features are important for predicting emotional states. However each state is characterized by its own set of relevant in-air and on-paper features and tasks. In particular, depression is characterized by cues extracted from drawing tasks while anxiety and stress use both writing and drawing cues. Anxiety and stress were found more difficult to recognize compared to depression in terms of accuracy. The present research involved a specific class of subjects (students). This was done on purpose in order to reduce the variability in the collected data. However, this is also a limitation since it does not allow for a generalization of the identified features and the obtained results to a larger sample of individuals. In the future we plan to analyze other population groups such as kids, mature and elder people, looking for commonalities and differences in order to delineate the proposed procedure and identify a general methodology for detecting emotional disorders from handwriting data. We consider that the first step must be a study in each group because a lot of indicators are related to age (for instance, it is believed that happiness in function or years has a U shape with a minimum around 45 years old). Nevertheless, the strengths of the proposed work are:

- Proposing, for the first time, the exploitation of handwriting data for the detection of emotional disorders
- Proposing and experimental design for the data collection.
- Proposing an automatic procedure for the selection of appropriate handwriting features useful for the task at the hand; these features are easy to collect and do not need manual measurements.
- Providing a handwriting database to the scientific community for further testing.
- Opening new investigation approaches to help psychologists and clinicians to assess emotional disorders.

The set of features we have proposed to extract is restricted to timing and ductus-based features. The stress state, which induces modifications in the muscle tonus may benefit from new features linked to pressure [14]. Higher level features such as fluidity of movements could be added to this set by an analysis of speed changes. Model-based features could also be extracted such as the parameters provided by the kinematic theory of rapid human movements [26]. Thus we hope that the database release will trigger proposals for new features and recognition approaches, as well as proposals for extension.

**Acknowledgments**

This work has been supported by FEDER and Ministerio de ciencia e Innovacin, TEC2012-38630-C04-03.